\documentclass[a4paper, 10pt, conference]{ieeeconf}
\IEEEoverridecommandlockouts
\pdfoutput=1

\usepackage{amsmath,amssymb,amsfonts}
\usepackage{graphicx}
\usepackage{textcomp}
\usepackage{xcolor}
\usepackage{tikz}
\usepackage{pgfplots}
\usetikzlibrary{angles,quotes}
\pgfplotsset{compat=newest}
\pgfplotsset{every axis legend/.append style={%
		cells={anchor=west}}
}
\pgfplotsset{every y tick label/.append style={font=\footnotesize}}
\pgfplotsset{every x tick label/.append style={font=\footnotesize}}
\pgfplotsset{every axis x label/.append style={font=\footnotesize}}
\pgfplotsset{every axis y label/.append style={font=\footnotesize}}
\pgfplotsset{every axis legend/.append style={font=\footnotesize}}
\pgfplotsset{every axis title/.append style={font=\footnotesize}}
\usepgfplotslibrary{polar}
\usetikzlibrary{arrows}
\tikzset{>=stealth'}
\usepgfplotslibrary{groupplots}
\usepackage[per-mode=symbol,detect-all]{siunitx}
\DeclareSIUnit\feet{ft}
\DeclareGraphicsExtensions{.pdf, .jpg, .png}
\usepackage[noend]{algorithmic}
\usepackage{algorithm}
\usepackage{array}
\usepackage{booktabs}
\usepackage[keeplastbox]{flushend}
\usepackage[capitalize]{cleveref}
\usepackage{todonotes}

\usepackage[backend=bibtex,style=ieee,maxcitenames=1,mincitenames=1]{biblatex}
\newcommand{\ubar}[1]{\text{\b{$#1$}}}

\AtEveryBibitem{
	\ifentrytype{inproceedings}{
		\clearlist{address}
		\clearlist{publisher}
		\clearname{editor}
		\clearlist{organization}
		\clearfield{url}  
		\clearfield{url}  
		\clearfield{doi}  
		\clearfield{pages}  
		\clearlist{location}}{}}
\AtEveryBibitem{
	\clearfield{doi}
	\clearfield{issn}
	\clearfield{month}  
}
\AtEveryBibitem{
	\ifentrytype{article}{
		\clearlist{address}
		\clearlist{publisher}
		\clearname{editor}
		\clearlist{organization}
		\clearfield{url}  
		\clearfield{doi}  
		\clearlist{location}}{}}
    
\begin{document}
\title{\LARGE \bf Validation of Image-Based Neural Network Controllers \\ through Adaptive Stress Testing}
\author{Kyle D. Julian$^1$, Ritchie Lee$^2$, and Mykel J. Kochenderfer$^1$%
\thanks{This material is based upon work supported by the National Science Foundation Graduate Research Fellowship under Grant No. DGE-1656518, AFRL and DARPA under contract
FA8750-18-C-0099, and a NASA Ames Research Center summer internship. Any opinions, findings, or recommendations expressed in this material are those of the authors and do not necessarily reflect the views of the U.S. Government.}
\thanks{$^1$Kyle D. Julian and Mykel J. Kochenderfer are with Aeronautics and
Astronautics, Stanford University, Stanford, CA 94305, USA {\tt\small \{kjulian3,
mykel\}@stanford.edu}}
\thanks{$^2$Ritchie Lee is with KBR Inc. at NASA Ames Research Park, Moffett Field, CA 94035, USA {\tt\small ritchie.lee@nasa.gov}}
}
\maketitle

\begin{abstract}
Neural networks have become state-of-the-art for computer vision problems because of their ability to efficiently model complex functions from large amounts of data. While neural networks can be shown to perform well empirically for a variety of tasks, their performance is difficult to guarantee. Neural network verification tools have been developed that can certify robustness with respect to a given input image; however, for neural network systems used in closed-loop controllers, robustness with respect to individual images does not address multi-step properties of the neural network controller and its environment. Furthermore, neural network systems interacting in the physical world and using natural images are operating in a black-box environment, making formal verification intractable. This work combines the adaptive stress testing (AST) framework with neural network verification tools to search for the most likely sequence of image disturbances that cause the neural network controlled system to reach a failure. An autonomous aircraft taxi application is presented, and results show that the AST method finds failures with more likely image disturbances than baseline methods. Further analysis of AST results revealed an explainable cause of the failure, giving insight into the problematic scenarios that should be addressed. 
\end{abstract}

\section{Introduction}
Many autonomous systems interact in complex environments and operate with high-dimensional data, such as images. Recent work has shown that neural networks can be trained efficiently to make decisions for image-based problems. \citeauthor{mnih2015human} use deep reinforcement learning to train neural network controllers that map Atari screen images to controller commands to create game-playing agents that outperform humans~\cite{mnih2015human}. Additional work has shown that neural networks can play chess~\cite{silver2018general}, classify objects~\cite{krizhevsky2012imagenet}, and recognize digits~\cite{lecun1998gradient}. In addition, neural networks can be used to control vehicles, such as steering cars~\cite{pomerleau1989alvinn,bojarski2016end}, guiding aircraft to waypoints~\cite{julian2017neural}, and controlling quadrotors~\cite{bansal2016learning}.

Although misclassifying a cat image is undesirable, steering vehicles off roads or into other vehicles can be catastrophic. Recently, tools have been developed that can verify input-output properties of neural networks, such as those representing aircraft collision avoidance policies~\cite{katz2017reluplex}. These tools use the simplex method~\cite{katz2017reluplex,katz2019marabou}, mixed integer linear programming~\cite{lomuscio2017approach}, symbolic interval analysis with linear relaxation~\cite{wang2018efficient}, and other approaches~\cite{liu2019algorithms}.

However, verifying input-output properties of systems acting as closed-loop controllers is insufficient to verify safety. Additional work has focused on verification of multi-step properties of neural network controllers acting within their environment~\cite{ivanov2019verisig,akintunde2018reachability,xiang2018reachability,julian2019guaranteeing}. Existing closed-loop verification work is applicable when the network input is low-dimensional and adequate environmental models exist; however, when the network input is high-dimensional and the environment is complex, verification approaches are intractable. As a result, many image-based neural network verification approaches focus on local robustness around validation images~\cite{katz2017towards,gopinath2018deepsafe}. However, local robustness is an input-output property and does not address closed-loop safety.

This work focuses on validation of image-based neural network controllers. Existing work on validation of complex systems has led to the development of adaptive stress testing (AST), which uses reinforcement learning to find the most likely ways systems fail~\cite{lee2015adaptive,koren2019efficient}. However, existing work with AST has only considered low-dimensional problems.

Our method combines ideas from local robustness verification with AST black box validation to efficiently search for sequences of image disturbances that lead to failure. Using neural network verification tools allows the algorithm to search for multi-step sequences in a lower-dimensional space than the size of the image. As a result, this method scales well to neural networks with hundreds of input variables without making any assumptions about the environment, allowing the tool to be easily integrated with any existing simulator. An example aircraft taxiway application is presented that uses the X-Plane 11 photo-realistic flight simulator~\cite{xplane11}. The method is able to find sequences of image disturbances that cause the neural network to guide the aircraft off the taxiway, and further analysis reveals explainable weaknesses of the neural network that were exploited to cause failures. 
\section{Background}
This work combines ideas from reinforcement learning and neural network verification, which are described in this section.

\subsection{Markov Decision Process}
A Markov decision process (MDP) is a general framework for modeling sequential decision-making problems and is described by the tuple $(\mathcal{S},\mathcal{A},R,T)$~\cite{Kochenderfer2015chapter4}. An agent in state $s \in \mathcal{S}$ takes action $a \in A$, transitions to $s'$ with probability $T(s'\mid s,a)$, and receives reward $r=R(s,a,s')$. The action-value function $Q^\pi(s,a)$ gives the expected value of taking action $a$ from state $s$ and following the policy $a=\pi(s)$ for all future states, computed as 

\begin{equation}
    Q^\pi(s,a) = \mathbb{E}\left[ \sum_{t=0}\gamma^t r_t \mid s_0=s, a_0=a, a'=\pi(s) \right],\label{eq:rl}
\end{equation}
where discount factor $\gamma$ is set to $\gamma=1$ for finite horizon problems and $0<
\gamma<1$ for infinite horizon problems. The goal with an MDP to compute $\pi(s)$ that maximizes $Q(s,a)$, denoted as $\pi^*(s)$ and $Q^*(s,a)$ respectively.
 
If the transition function is known, then optimization algorithms can compute $\pi^*(s)$; however, when the transition function is unknown or difficult to model, such as with image-based navigation, alternative methods are needed. Reinforcement learning is a model-free method that uses simulations to estimate $Q^*(s,a)$. The following two subsections describe reinforcement learning algorithms used in this work.

\subsection{Monte Carlo Tree Search}
Monte Carlo tree search (MCTS) begins with a root node with initial state, $s_0$. New states $s'$ are added to the tree as leaf nodes from an existing node $s$ via the action $a$ used to arrive at $s'$ from $s$. At each node, the algorithm keeps track of $N(s)$, the number of times the node with state $s$ has been visited, $N(s,a)$, the number of times action $a \in A(s)$ has been taken from $s$, and $Q(s,a)$, the value estimate of taking action $a$ from state $s$. MCTS follows three basic steps:
\begin{enumerate}
    \item \emph{Search}. Beginning at the root node with state $s$ and actions $A(s)$ already added to the tree, new actions are added to the tree if $\Vert A(s) \Vert < kN(s)^\alpha$, where $k$ and $\alpha$ are hyperparameters that balance exploration with exploitation. If a new action is not added, then an existing action is taken to maximize
    \begin{equation}
        Q(s,a)+c\sqrt{\frac{\log N(s)}{N(s,a)}}
    \end{equation}
    where $c$ is also a hyperparameter. The search process repeats from $s'$ until a new action is selected.
    
    \item \emph{Expansion}. The tree is expanded with the new action by simulating the action to compute $s'$, which is added as a new leaf node to the tree with an empty $A(s')$. This work assumes that transitions are deterministic, although other versions of MCTS can incorporate stochastic actions~\cite{lee2015adaptive}.
    
    \item \emph{Rollout}. Once a new state $s'$ has been added to the tree, a random rollout simulation is used to initialize $Q(s,a)$. The rollout uses a default policy $\pi_0(s)$ to determine actions taken, and the rollout continues until a pre-determined depth or terminal state is reached. $Q(s,a)$ is computed from the rollout simulation using \cref{eq:rl} with $\gamma=1$ since the rollout has a finite depth. The estimated $Q(s,a)$ is then propagated up through all parent nodes according to 
    \begin{align}
        N(s,a) &\leftarrow N(s,a) + 1 \\
        q &\leftarrow R(s,a,s') + \gamma Q(s',a') \\
        Q(s,a) &\leftarrow Q(s,a) + \frac{q-Q(s,a)}{N(s,a)}
    \end{align}
    where $Q(s',a')$ is the updated value of the child node.
\end{enumerate} 
Once the rollout finishes, MCTS begins another iteration by searching from the root node until a new state is added. For further details on the MCTS algorithm, see~\cite{lee2015adaptive}.

\subsection{Deep Q-Learning}
While MCTS can simulate many trajectories to optimize $\pi(s)$, the algorithm does not generalize from the values of states already seen to predict values of new states. Another popular algorithm, deep Q-learning, or deep Q-networks (DQN), maintains a global functional representation of $Q(s,a)$~\cite{mnih2015human}. As a result, updating $Q(s,a)$ for one state updates the values for nearby states.

DQN uses a deep neural network to approximate $Q(s,a)$, and the network parameters are updated through gradient descent methods using a loss value based on the temporal difference between $r+\gamma Q(s',a')$ and $Q(s,a)$. This work used the OpenAI Baselines implementation of DQN~\cite{baselines} with prioritized experience replay~\cite{schaul2015prioritized,baselines}.

\subsection{Neural Network Verification}\label{subsec:nnv}
Recent advancements have produced tools that verify neural network input-output properties~\cite{katz2017reluplex,katz2019marabou,wang2018efficient}. For neural networks of the form $y=f(x)$, where $f$ is composed of computational layers with piecewise-linear activations, these properties are defined as $x \in \mathcal{X} \implies f(x) \notin \mathcal{Y}$, where $\mathcal{X}$ and $\mathcal{Y}$ are convex polytopes. The verification tools  provide either a guarantee that the property holds (UNSAT) or a satisfying $x'$ such that $x' \in \mathcal{X} \land f(x) \in \mathcal{Y}$ (SAT).

When $x$ is an image, neural network verification tools can compute the robustness of the neural network to noise added to a given image. Local robustness around an image $x$ for a neural network with a scalar output can be defined as
\begin{equation}
    \Vert \tilde{x} \Vert_\infty < \delta \implies \vert f(x+\tilde{x}) - f(x) \vert < \epsilon
\end{equation}
where $\delta$ limits changes to pixels, $\epsilon$ limits change to the network output, and $\tilde{x}$ is image noise. Verifying robustness around validation images can check if the network is overly sensitive to small perturbations, but robustness alone cannot determine what level of perturbations can be safely tolerated.
\section{Methodology} %
This section describes the Adaptive Stress Testing method used to validate image-based neural network controllers.

\subsection{Adaptive Stress Testing}
Adaptive Stress Testing (AST), is a particular configuration of model-free reinforcement learning (RL). Rather than learning a policy that optimizes performance of an agent in an environment, AST optimizes the environment to cause a learned agent to fail. AST uses states $s$ that define the state of the simulator and actions $a$ that define environmental factors controlled by the simulator. This work considers image-based neural networks, so actions are disturbances to the input images of the controller.

AST treats the simulator as a black box and interacts with the simulator through the following functions:
\begin{enumerate}
    \item $\text{Initialize}(s)$. Load a state of the simulator.
    \item $\text{Step}(s,a)$. Advance the simulator one step from state $s$ given action $a$.
    \item $\text{IsTerminal}(s)$. Return true if state is a terminal state.
    \item $\text{IsFailure}(s)$. Return true if the state is a failure state.
\end{enumerate}
The AST reward function $R(s,a)$ encourages RL algorithms to find the most likely sequence of actions that causes the system to fail, as discussed in the following subsection.

\subsection{Reward Function}
The goal of AST is to find the most likely sequence of actions $a_{0:t-1}$ from $s_0$ with $s_i=\text{Step}(s_{i-1},a_{i-1})$ such that $\text{IsFailure}(s_t)$ is true. Assuming that each action is independent, the optimization problem becomes 
\begin{align}
    \underset{a_{0:t-1}}{\text{maximize}} &\prod_{i=0}^{t-1} p(a_i) \\
    \text{subject to}\ \ &\text{IsFailure}(s_t)  \nonumber
\end{align}
where $p(a_i)$ is the likelihood of the environment producing action $a_i$.

To incentivize RL algorithms to find such a sequence of actions, AST uses reward function $R(s,a)$ defined as 
\begin{equation}
    R(s,a) = \begin{cases} 
    0, &\mbox{if } \text{IsFailure(s)} \\
    \log p(a), &\mbox{else if not } \text{IsTerminal(s)} \\
    -\alpha -\beta \times \text{Dist}(s), &\mbox{otherwise}
    \end{cases}
\end{equation}
where \text{Dist}(s) is some measure of the simulator's closeness to a failure, and $\alpha$ and $\beta$ scale the penalty term given when a terminal state is reached that is not a failure~\cite{koren2019efficient}. In practice $\alpha$ and $\beta$ are very large to encourage the simulator to find a failure before optimizing the action sequence to reach failure. The following subsection further describes the action space when testing image-based neural networks.

\begin{figure*}
    \centering
    \includegraphics[width=0.28\linewidth]{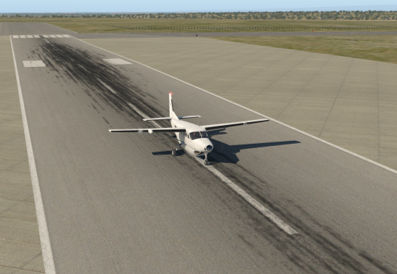}
    \hspace{0.02\linewidth}
    \includegraphics[width=0.35\linewidth]{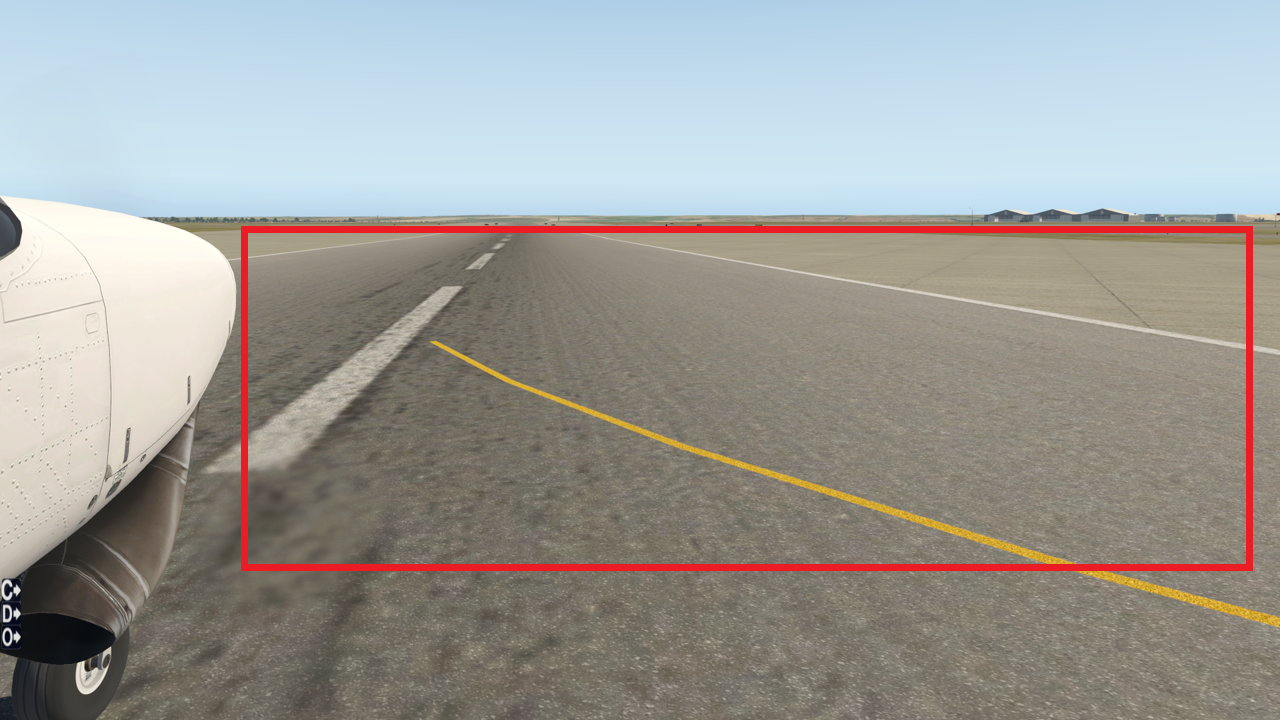}
    \hspace{0.02\linewidth}
    \includegraphics[width=0.27\linewidth]{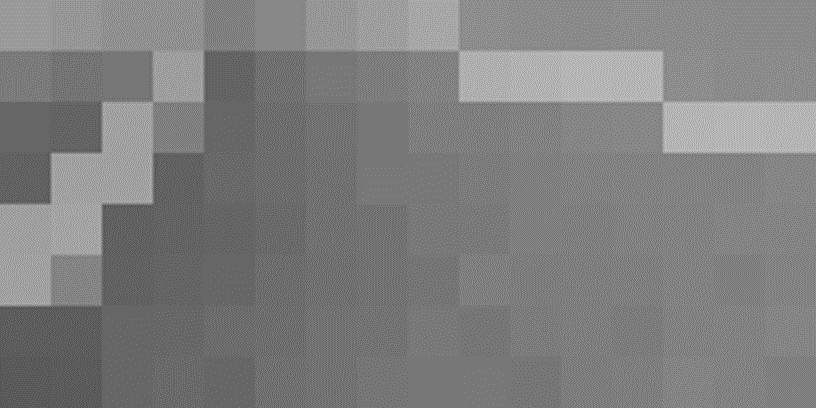}
    \caption{X-Plane 11 aircraft on taxiway (left), view from camera with cropped region shown in red (middle), and downsampled taxiway image (right)}
    \label{fig:aircraft}
\end{figure*}

\subsection{Action Space}

\begin{algorithm}[tb]
	\caption{Parallel image disturbance optimization}
	\label{alg:algorithm}
	\textbf{Input}: $f$, $x$, $\delta$, $N$, $\text{tol}$, $\bar{\epsilon}$ \\
	\textbf{Output}: $\ubar{\epsilon},\ \tilde{x}$
	\begin{algorithmic}[1] %
	    \STATE $\ubar{\epsilon},\ \tilde{x} = 0$
		\WHILE{$\bar{\epsilon} - \ubar{\epsilon}>\text{tol}$}
		\FOR{$i = 1:N$}
		\STATE $\epsilon_i = \ubar{\epsilon} + (\bar{\epsilon}-\ubar{\epsilon})\times i/(N+1)$
		\STATE $\text{result}_i,\ \tilde{x}_i=\text{solve}(f,x,\delta,\epsilon_i)$
		\ENDFOR
		\STATE Run all solve calls in parallel
		\STATE $\text{result}_0,\ \tilde{x}_0,\ \epsilon_0 = \text{SAT},\ \tilde{x},\ \ubar{\epsilon}$
		\STATE $\text{result}_{N+1},\ \epsilon_{N+1}  = \text{UNSAT},\ \bar{\epsilon}$
		\STATE $\ubar{\epsilon} \leftarrow \max \epsilon_i \text{ s.t. } \text{result}_i = \text{SAT}$
		\STATE $\bar{\epsilon} \leftarrow \min \epsilon_i \text{ s.t. } \text{result}_i = \text{UNSAT}$
		\STATE $\tilde{x} \leftarrow \tilde{x}_i \text{ s.t. } \epsilon_i = \ubar{\epsilon}$
		\ENDWHILE
		\STATE \textbf{return} $\ubar{\epsilon},\ \tilde{x}$
	\end{algorithmic}
\end{algorithm}

In previous work with AST, the action space has been low-dimensional~\cite{lee2015adaptive,koren2019efficient}. However, when the action space is a high-dimensional image, previous AST algorithms will not perform well. The size of the action space grows exponentially with the dimensionality, so AST would need to sample exponentially more actions to achieve good performance.

This work presents a different approach to computing actions that scales better to high-dimensional spaces. Given a neural network controller $f$ that maps images $x$ to a scalar value $y$, neural network verification tools can compute an image disturbance $\tilde{x}$ that changes the network output as much as possible assuming that $x+\tilde{x}$ is in the neighborhood of a given image $x$. This work defines the neighborhood around $x$ as images where each pixel changes by at most $\delta$, though other definitions could be used. For a given input image $x$, the problem becomes
\begin{align} 
    \underset{\tilde{x}}{\text{maximize }} &f(x+\tilde{x}) - f(x) \label{eq:optAction} \\ 
    \text{subject to}\ \ & \Vert \tilde{x} \Vert_\infty \leq \delta. \nonumber
\end{align}
\Cref{eq:optAction} can also be written with minimize to compute the perturbation that minimizes the neural network output.

As described in \Cref{subsec:nnv}, neural network verification tools can compute local robustness properties and verify that the output does not change by more than $\epsilon$ (UNSAT) or provide a counterexample where $\tilde{x}$ satisfies $\Vert \tilde{x} \Vert_\infty \leq \delta$ and $f(x+\tilde{x}) - f(x) \ge \epsilon$ (SAT). To use these tools in an optimization problem as described in \cref{eq:optAction} rather than a satisfiability problem, multiple queries need to be evaluated to search for the largest $\epsilon$ value that returns SAT. This search can be done in parallel using the algorithm described in \Cref{alg:algorithm}, which requires the number of parallel queries to run, $N$, $\epsilon$ tolerance, $\text{tol}$, and an upper bound on $\epsilon$, $\bar{\epsilon}$. If no upper bound is known ahead of time, then the algorithm can be modified to increase $\bar{\epsilon}$ until the solver returns UNSAT. \Cref{alg:algorithm} describes the parallel search for the largest positive change to the network output; the largest negative change to the network output can also be computed in a similar manner. For simplicity, negative values of $\delta$ and $\epsilon$ used in this work imply a search for the disturbance $\tilde{x}$ that minimizes network output.

\Cref{alg:algorithm} returns the largest $\epsilon$ for which there exists a satisfying $\tilde{x}$. Using this approach, the AST action $a$ is equivalent to the image disturbance $\tilde{x}$, effectively reducing the action space for AST from the dimensionality of $\tilde{x}$ to one dimension, $\delta$. Furthermore, since $a$ is constrained by $\Vert a \Vert_\infty \le \delta$, $p(a)$ can be defined as a function of $\delta$ using
\begin{align}
    p(a) &= \mathcal{N}(\Vert a \Vert_\infty \mid 0, \sigma^2) \\
    &= \mathcal{N}(\delta \mid 0, \sigma^2)  \nonumber
\end{align}
where $\sigma$ defines a zero-mean univariate normal distribution. This approach implies that small image disturbances are more likely than large disturbances. Using this approach, AST can be applied to image-based control problems.
\section{Aircraft Taxi Application}
An aircraft taxi problem is presented here to demonstrate the neural network adaptive stress testing method. A Cessna 208B Grand Caravan simulated in X-Plane 11~\cite{xplane11} is taxiing at \SI{5}{\meter\per\second} along runway 04 of Grant County International Airport and must stay on the taxiway using only images taken once per second from a camera on the right wing of the aircraft. A neural network is trained through supervised learning to map runway images to crosstrack position $d$ and heading angle $\theta$, which are used to control the aircraft. The taxiway center is defined as $d=\SI{0}{\meter}$, and the taxiway heading angle is defined as $\theta=\ang{0}$. The following subsections discuss the design of the neural network controller, preliminary validation, and AST experimental setup.

\subsection{Neural Network Design}
Existing works show that image-based neural networks are susceptible to adversarial attacks, where small changes to pixel values cause large changes to the network output~\cite{goodfellow2014explaining,carlini2017towards}. Because this application uses runway images as inputs, the images can be downsampled significantly without losing important information, which allows the trained network to be smaller as well. The reduced input representation and network size may help the network be more robust to pixel perturbations, and a smaller neural network representation will be more quickly verified by Marabou.

The following procedure was used to shrink input images from $200\times 360$ RGB images to $8\times 16$ grayscale images:
\begin{enumerate}
    \item Crop out the sky and airplane nose.
    \item Resize image to $128\times 256$ and convert to grayscale.
    \item Downsample image by splitting image into 128 $16\times 16$ boxes and averaging the 16 brightest pixels within each box, resulting in an $8\times 16$ image.
    \item Bias all pixel values so that the average value is 0.5 (when pixel values range from 0 to 1).
\end{enumerate}
Forcing the mean pixel value to be 0.5 helps the network generalize to different lighting conditions and increases robustness by adding a constraint to adversarial images.

A neural network with 3 hidden layers and 32 total ReLU activations was trained. The network is composed of an $8\times 8$ convolutional layer with 8 filters and stride of 8 followed by two fully connected layers of size 8. The architecture was designed to contain as few ReLUs as possible while still providing accurate estimates of $d$ and $\theta$. Network outputs $d$ and $\theta$ are combined into a rudder command $r$ with proportional control law
\begin{equation}
    r = 0.015d + 0.008\theta.
\end{equation}
Because the final layer of the neural network and proportional control law are linear, they can be combined by modifying the last layer of the neural network. The final neural network controller maps $8\times 16$ downsampled images of the runway to rudder commands.

\subsection{Preliminary Validation}
To test the controller, 100 random simulations were run for four cloud conditions with initial crosstrack position $d_0\sim\mathcal{U}(\SI{-5}{\meter},\SI{5}{\meter})$ and heading angle $\theta_0 \sim \mathcal{U}(\ang{-20},\ang{20})$ at a random simulated time between 9am and 3pm. After 20 seconds elapsed, crosstrack position had an average absolute value of \SI{0.287}{\meter} with standard deviation \SI{0.534}{\meter} and maximum value \SI{1.015}{\meter}, and the heading angle had an absolute value under \ang{2} for all 400 simulations. 

These results suggest that the controller performs well, but success in nominal conditions does not mean the controller will always perform correctly. Taxiways in the real world have skid marks, reflective puddles, paint spots, and more, so the controller also needs to perform well when the images are perturbed. The method proposed here addresses these questions to better understand the network shortcomings and failure modes, as discussed in \Cref{sec:Results}.

\subsection{Experimental Setup}

The Marabou tool was used to generate image perturbations because the tool scales well to high-dimensional inputs and the python interface was easy to integrate to other python components~\cite{katz2019marabou}. Three types of AST experiments were run: MCTS with Marabou, DQN with Marabou, and MCTS with random samples. All experiments used a discrete number of actions, $N_\text{actions}$, with $\delta \in \text{linspace}(-\delta_\text{max},\delta_\text{max},N_\text{actions})$. MCTS begins with a root node near the center of the taxiway, while the initial state for DQN trajectories is initialized with $d_0\sim\mathcal{U}(\SI{-5}{\meter},\SI{5}{\meter})$ and $\theta_0 \sim \mathcal{U}(\ang{-20},\ang{20})$. When random samples are used instead of Marabou, 5000 random image perturbations are sampled, and the perturbation that changes the network output the most is used. The time required to sample and evaluate 5000 images is approximately the time used by Marabou. Failure states are defined as $\vert d \vert > \SI{10}{\meter}$, and terminal states are defined as more than $\SI{200}{\meter}$ downtrack.

The experiments use NASA's open source XPlaneConnect to interface with X-Plane 11. Experiments were conducted on a desktop with 16GB of RAM, a 6 core Intel i7 processor, and an NVidia GeForce GTX 1070 Ti GPU. Ten Marabou queries were run in parallel with a rudder disturbance tolerance of 0.003. To speed up MCTS rollouts, a set of rollouts were pre-computed and used to approximate new rollout values through linear interpolation.
\section{Results} \label{sec:Results}

\begin{table}
    \centering
    \caption{AST results using MCTS}
    \begin{tabular}{lccccc}  
    	\toprule
    	$N_\text{actions}$  & $\delta_\text{max}$ & Failure? & Steps & $\delta_\text{mean}$ & Log-likelihood \\
    	\midrule
    	2 & 0.02 & No & N/A & N/A & N/A \\
    	2 & 0.027 & No & N/A & N/A & N/A \\
    	2 & 0.028 & Yes & 26 & 0.028 & $-125.812$ \\
    	2 & 0.029 & Yes & 24 & 0.029 & $-122.975$ \\
    	2 & 0.035 & Yes & 21 & 0.035 & $-147.923$ \\
    	2 & 0.040 & Yes & 11 & 0.040 & $-98.108$ \\
    	3 & 0.035 & Yes & 35 & 0.030 & $-215.913$\\
    	4 & 0.035 & Yes & 35 & 0.031 & $-213.871$ \\
    	6 & 0.035 & Yes & 46 & 0.028 & $-249.541$ \\
    	\bottomrule
    \end{tabular}
    \label{tab:mcts}
\end{table}

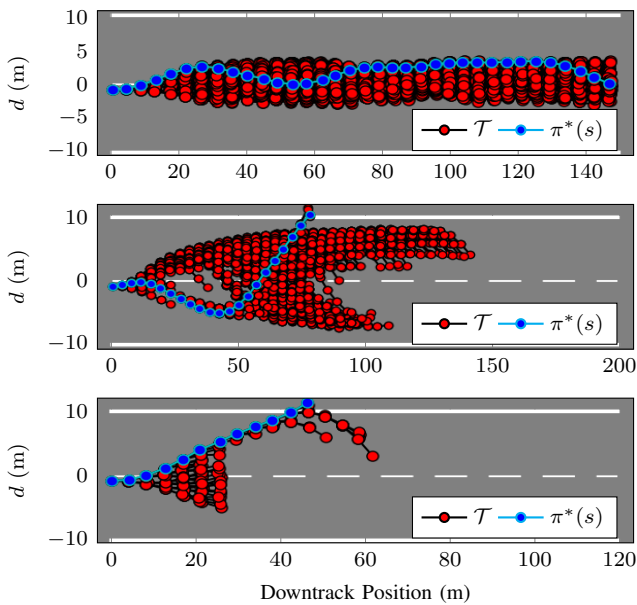
\begin{figure}
    \centering
    \begin{tikzpicture}[]
\begin{groupplot}[height={3.5cm},width={\linewidth},group style={horizontal sep=0.0cm,vertical sep=0.65cm, group size=1 by 3}]

\nextgroupplot [ylabel={$d\ (\si{\meter})$}, enlargelimits = false, axis on top,xmin=-4.0,xmax=154,ymin=-10.7,ymax=11.0,legend pos={south east},
legend columns=2]]
\addplot+ [mark=*,black,thick,mark options={mark size=1.7,fill=red}]coordinates {
(-100.0, 0.0)
};
\addlegendentry{$\mathcal{T}$}
\addplot+ [mark=*,cyan,thick,mark options={mark size=1.7,fill=blue}]coordinates {
(-100.0, 0.0)
};
\addlegendentry{$\pi^*(s)$}

\addplot graphics
[xmin=-4,xmax=154,ymin=-10.7,ymax=11.0]
{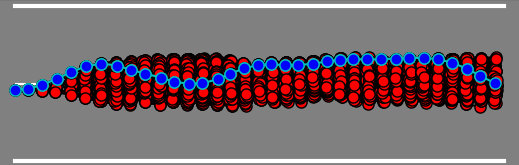};

\nextgroupplot [ylabel={$d\ (\si{\meter})$}, enlargelimits = false, axis on top,xmin=-5.5,xmax=205.5,ymin=-11,ymax=12,legend pos={south east},
legend columns=2]]
\addplot+ [mark=*,black,thick,mark options={mark size=1.7,fill=red}]coordinates {
(-100.0, 0.0)
};
\addlegendentry{$\mathcal{T}$}
\addplot+ [mark=*,cyan,thick,mark options={mark size=1.7,fill=blue}]coordinates {
(-100.0, 0.0)
};
\addlegendentry{$\pi^*(s)$}

\addplot graphics
[xmin=-5.5,xmax=205.5,ymin=-11,ymax=12]
{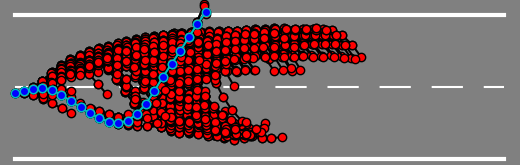};

\nextgroupplot [ylabel={$d\ (\si{\meter})$},xlabel={Downtrack Position (m)}, enlargelimits = false, axis on top,xmin=-3.3,xmax=123.3,ymin=-10.5,ymax=12,legend pos={south east},
legend columns=2]]
\addplot+ [mark=*,black,thick,mark options={mark size=1.7,fill=red}]coordinates {
(-100.0, 0.0)
};
\addlegendentry{$\mathcal{T}$}
\addplot+ [mark=*,cyan,thick,mark options={mark size=1.7,fill=blue}]coordinates {
(-100.0, 0.0)
};
\addlegendentry{$\pi^*(s)$}

\addplot graphics
[xmin=-3.3,xmax=123.3,ymin=-10.5,ymax=12]
{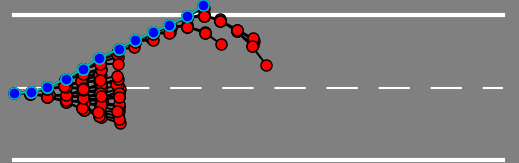};

\end{groupplot}

\end{tikzpicture}
    \caption{MCTS search trees when using two actions with $\delta_\text{max}=0.02$ (top), $\delta_\text{max}=0.035$ (middle), and $\delta_\text{max}=0.04$ (bottom)}
    \label{fig:MCTS_Tree_2}
\end{figure}

A summary of results using MCTS with Marabou is shown in \Cref{tab:mcts}. When only two actions are used, $\delta$ is always equal to $\delta_\text{max}$, while larger numbers of discrete actions also use actions with $\delta$ values less than the maximum. When $\delta_\text{max}$ is too small, AST cannot find a sequence of image disturbances that leads to failure, while failure sequences are easily found for large $\delta_\text{max}$. At some critical $\delta_\text{max}$, image disturbances are just strong enough to control the aircraft off the taxiway. Increasing the number of actions makes finding a failure sequence more difficult because MCTS also considers many weak image disturbances, which are less likely to result in failure. As a result, MCTS finds longer sequences to failure, which has a lower log-likelihood but also a lower $\delta_\text{mean}$.

\Cref{fig:MCTS_Tree_2} shows the MCTS search tree $\mathcal{T}$ along with optimized policy $\pi^*(s)$. For $\delta_\text{max}=0.02$ the aircraft never exceeds $\SI{5}{\meter}$ from the centerline. For $\delta_\text{max}=0.04$, disturbances that always maximize the neural network output can push the aircraft off the left side of the taxiway. For $\delta_\text{max}=0.035$, always maximizing the neural network output will not be enough to leave the taxiway. However, AST is able to find a failure by first decreasing the network output to turn the aircraft right before increasing the network output and turning the aircraft sharply left and off the taxiway. This failure sequence represents a novel failure mode that is non-obvious.

\begin{figure}
    \centering
    \begin{tikzpicture}[]
\begin{groupplot}[height={3.5cm},width={\linewidth},group style={horizontal sep=0.0cm,vertical sep=1.0cm, group size=1 by 1}]

\nextgroupplot [ylabel={$d\ (\si{\meter})$},xlabel={Downtrack Position (m)}, enlargelimits = false, axis on top,xmin=-5.5,xmax=205.5,ymin=-10.7,ymax=13,legend pos={north west},
legend columns=2]]
\addplot+ [mark=*,black,thick,mark options={mark size=1.7,fill=red}]coordinates {
(-100.0, 0.0)
};
\addlegendentry{$\mathcal{T}$}
\addplot+ [mark=*,cyan,thick,mark options={mark size=1.7,fill=blue}]coordinates {
(-100.0, 0.0)
};
\addlegendentry{$\pi^*(s)$}

\addplot graphics
[xmin=-5.5,xmax=205.5,ymin=-10.7,ymax=13]
{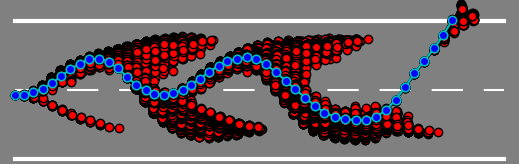};

\end{groupplot}

\end{tikzpicture}
    \caption{MCTS search tree when using six actions with $\delta_\text{max}=0.035$}
    \label{fig:MCTS_Tree_6}
\end{figure}

\Cref{fig:MCTS_Tree_6} shows the search tree for MCTS using 6 discrete actions. The sequence found is longer than when using only two actions, but the behavior is similar. The image disturbances cause the aircraft to oscillate across the centerline like a pendulum, eventually gaining enough momentum to leave the taxiway.

\begin{figure}
    \centering
    \begin{tikzpicture}[]
\begin{groupplot}[height={3.5cm},width={\linewidth},group style={horizontal sep=0.0cm,vertical sep=1.0cm, group size=1 by 1}]

\nextgroupplot [ylabel={$d\ (\si{\meter})$},xlabel={Downtrack Position (m)}, enlargelimits = false, axis on top,xmin=-5.5,xmax=205.5,ymin=-10.7,ymax=13.5,legend pos={north west},
legend columns=2]]

\addplot graphics
[xmin=-5.5,xmax=205.5,ymin=-10.7,ymax=13.5]
{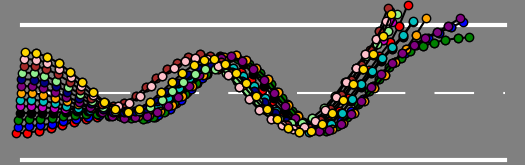};

\end{groupplot}

\end{tikzpicture}
    \caption{Simulated trajectories using learned DQN policy}
    \label{fig:dqn}
\end{figure}

\begin{figure*}
    \centering
    \begin{tikzpicture}[]

\begin{groupplot}[height=3.2cm, width=3.7cm, ytick style={draw=none},xtick style={draw=none},group style={horizontal sep=0.0cm,vertical sep=0.0cm, group size=8 by 4}]

\nextgroupplot [title style={align=center},
title = {$t=\SI{0}{\second}$\\$d=\SI{-0.74}{\meter}$\\$\theta=\SI{-0.10}{\degree}$\\$\epsilon=0.180$}, 
ylabel = {Original}, yticklabels={,,},xticklabels={,,}, enlargelimits = false, axis on top]
\addplot [point meta min=0, point meta max=50] graphics [xmin=-2.5, xmax=2.5, ymin=-1.0, ymax=1.0] {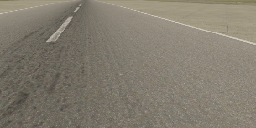};

\nextgroupplot [title style={align=center},
title = {$t=\SI{4}{\second}$\\$d=\SI{-0.41}{\meter}$\\$\theta=\SI{-5.80}{\degree}$\\$\epsilon=-0.104$}, 
ylabel = {}, yticklabels={,,},xticklabels={,,}, enlargelimits = false, axis on top]
\addplot [point meta min=0, point meta max=50] graphics [xmin=-2.5, xmax=2.5, ymin=-1.0, ymax=1.0] {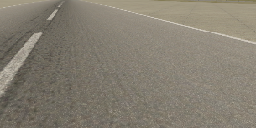};

\nextgroupplot [title style={align=center},
title = {$t=\SI{8}{\second}$\\$d=\SI{-3.51}{\meter}$\\$\theta=\SI{-9.78}{\degree}$\\$\epsilon=-0.071$}, 
ylabel = {}, yticklabels={,,},xticklabels={,,}, enlargelimits = false, axis on top]
\addplot [point meta min=0, point meta max=50] graphics [xmin=-2.5, xmax=2.5, ymin=-1.0, ymax=1.0] {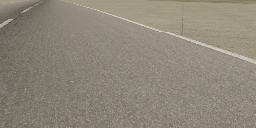};

\nextgroupplot [title style={align=center},
title = {$t=\SI{12}{\second}$\\$d=\SI{-4.70}{\meter}$\\$\theta=\SI{6.58}{\degree}$\\$\epsilon=0.163$}, 
ylabel = {}, yticklabels={,,},xticklabels={,,}, enlargelimits = false, axis on top]
\addplot [point meta min=0, point meta max=50] graphics [xmin=-2.5, xmax=2.5, ymin=-1.0, ymax=1.0] {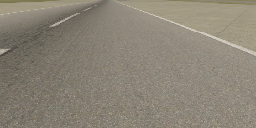};

\nextgroupplot [title style={align=center},
title = {$t=\SI{14}{\second}$\\$d=\SI{-3.88}{\meter}$\\$\theta=\SI{13.67}{\degree}$\\$\epsilon=0.195$}, 
ylabel = {}, yticklabels={,,},xticklabels={,,}, enlargelimits = false, axis on top]
\addplot [point meta min=0, point meta max=50] graphics [xmin=-2.5, xmax=2.5, ymin=-1.0, ymax=1.0] {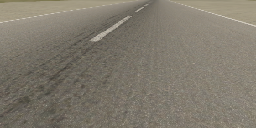};

\nextgroupplot [title style={align=center},
title = {$t=\SI{15}{\second}$\\$d=\SI{-2.55}{\meter}$\\$\theta=\SI{19.92}{\degree}$\\$\epsilon=0.188$}, 
ylabel = {}, yticklabels={,,},xticklabels={,,}, enlargelimits = false, axis on top]
\addplot [point meta min=0, point meta max=50] graphics [xmin=-2.5, xmax=2.5, ymin=-1.0, ymax=1.0] {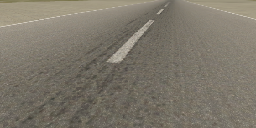};

\nextgroupplot [title style={align=center},
title = {$t=\SI{18}{\second}$\\$d=\SI{2.76}{\meter}$\\$\theta=\SI{24.44}{\degree}$\\$\epsilon=0.201$}, 
ylabel = {}, yticklabels={,,},xticklabels={,,}, enlargelimits = false, axis on top]
\addplot [point meta min=0, point meta max=50] graphics [xmin=-2.5, xmax=2.5, ymin=-1.0, ymax=1.0] {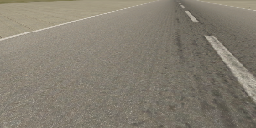};

\nextgroupplot [title style={align=center},
title = {$t=\SI{20}{\second}$\\$d=\SI{6.21}{\meter}$\\$\theta=\SI{23.75}{\degree}$\\$\epsilon=0.454$}, 
ylabel = {}, yticklabels={,,},xticklabels={,,}, enlargelimits = false, axis on top]
\addplot [point meta min=0, point meta max=50] graphics [xmin=-2.5, xmax=2.5, ymin=-1.0, ymax=1.0] {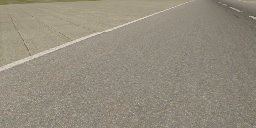};

\nextgroupplot [ylabel = {Downsampled}, yticklabels={,,},xticklabels={,,}, enlargelimits = false, axis on top]
\addplot [point meta min=0, point meta max=50] graphics [xmin=-2.5, xmax=2.5, ymin=-1.0, ymax=1.0] {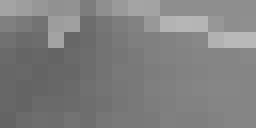};

\nextgroupplot [ylabel = {}, yticklabels={,,},xticklabels={,,}, enlargelimits = false, axis on top]
\addplot [point meta min=0, point meta max=50] graphics [xmin=-2.5, xmax=2.5, ymin=-1.0, ymax=1.0] {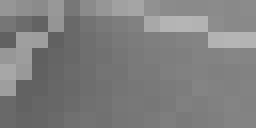};

\nextgroupplot [ylabel = {}, yticklabels={,,},xticklabels={,,}, enlargelimits = false, axis on top]
\addplot [point meta min=0, point meta max=50] graphics [xmin=-2.5, xmax=2.5, ymin=-1.0, ymax=1.0] {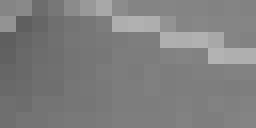};

\nextgroupplot [ylabel = {}, yticklabels={,,},xticklabels={,,}, enlargelimits = false, axis on top]
\addplot [point meta min=0, point meta max=50] graphics [xmin=-2.5, xmax=2.5, ymin=-1.0, ymax=1.0] {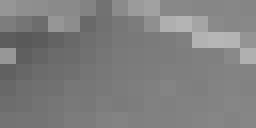};

\nextgroupplot [ylabel = {}, yticklabels={,,},xticklabels={,,}, enlargelimits = false, axis on top]
\addplot [point meta min=0, point meta max=50] graphics [xmin=-2.5, xmax=2.5, ymin=-1.0, ymax=1.0] {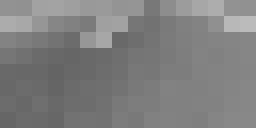};

\nextgroupplot [ylabel = {}, yticklabels={,,},xticklabels={,,}, enlargelimits = false, axis on top]
\addplot [point meta min=0, point meta max=50] graphics [xmin=-2.5, xmax=2.5, ymin=-1.0, ymax=1.0] {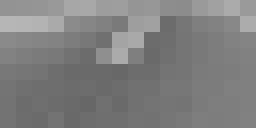};

\nextgroupplot [ylabel = {}, yticklabels={,,},xticklabels={,,}, enlargelimits = false, axis on top]
\addplot [point meta min=0, point meta max=50] graphics [xmin=-2.5, xmax=2.5, ymin=-1.0, ymax=1.0] {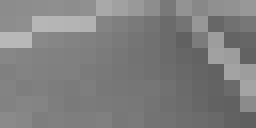};

\nextgroupplot [ylabel = {}, yticklabels={,,},xticklabels={,,}, enlargelimits = false, axis on top]
\addplot [point meta min=0, point meta max=50] graphics [xmin=-2.5, xmax=2.5, ymin=-1.0, ymax=1.0] {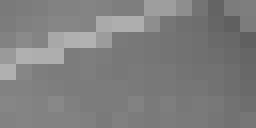};

\nextgroupplot [ylabel = {Perturbed}, yticklabels={,,},xticklabels={,,}, enlargelimits = false, axis on top]
\addplot [point meta min=0, point meta max=50] graphics [xmin=-2.5, xmax=2.5, ymin=-1.0, ymax=1.0] {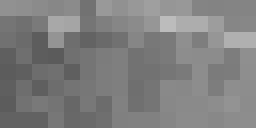};

\nextgroupplot [ylabel = {}, yticklabels={,,},xticklabels={,,}, enlargelimits = false, axis on top]
\addplot [point meta min=0, point meta max=50] graphics [xmin=-2.5, xmax=2.5, ymin=-1.0, ymax=1.0] {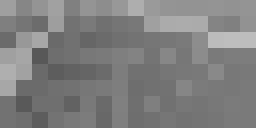};

\nextgroupplot [ylabel = {}, yticklabels={,,},xticklabels={,,}, enlargelimits = false, axis on top]
\addplot [point meta min=0, point meta max=50] graphics [xmin=-2.5, xmax=2.5, ymin=-1.0, ymax=1.0] {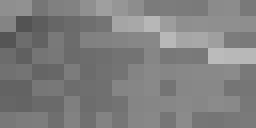};

\nextgroupplot [ylabel = {}, yticklabels={,,},xticklabels={,,}, enlargelimits = false, axis on top]
\addplot [point meta min=0, point meta max=50] graphics [xmin=-2.5, xmax=2.5, ymin=-1.0, ymax=1.0] {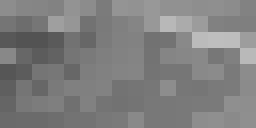};

\nextgroupplot [ylabel = {}, yticklabels={,,},xticklabels={,,}, enlargelimits = false, axis on top]
\addplot [point meta min=0, point meta max=50] graphics [xmin=-2.5, xmax=2.5, ymin=-1.0, ymax=1.0] {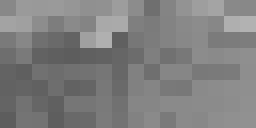};

\nextgroupplot [ylabel = {}, yticklabels={,,},xticklabels={,,}, enlargelimits = false, axis on top]
\addplot [point meta min=0, point meta max=50] graphics [xmin=-2.5, xmax=2.5, ymin=-1.0, ymax=1.0] {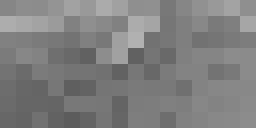};

\nextgroupplot [ylabel = {}, yticklabels={,,},xticklabels={,,}, enlargelimits = false, axis on top]
\addplot [point meta min=0, point meta max=50] graphics [xmin=-2.5, xmax=2.5, ymin=-1.0, ymax=1.0] {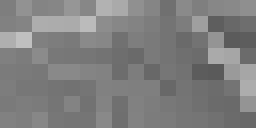};

\nextgroupplot [ylabel = {}, yticklabels={,,},xticklabels={,,}, enlargelimits = false, axis on top]
\addplot [point meta min=0, point meta max=50] graphics [xmin=-2.5, xmax=2.5, ymin=-1.0, ymax=1.0] {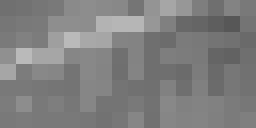};

\nextgroupplot [ylabel = {Reconstructed}, yticklabels={,,},xticklabels={,,}, enlargelimits = false, axis on top]
\addplot [point meta min=0, point meta max=50] graphics [xmin=-2.5, xmax=2.5, ymin=-1.0, ymax=1.0] {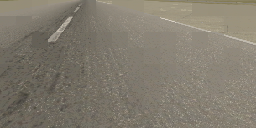};

\nextgroupplot [ylabel = {}, yticklabels={,,},xticklabels={,,}, enlargelimits = false, axis on top]
\addplot [point meta min=0, point meta max=50] graphics [xmin=-2.5, xmax=2.5, ymin=-1.0, ymax=1.0] {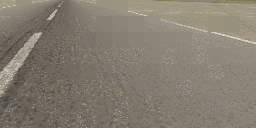};

\nextgroupplot [ylabel = {}, yticklabels={,,},xticklabels={,,}, enlargelimits = false, axis on top]
\addplot [point meta min=0, point meta max=50] graphics [xmin=-2.5, xmax=2.5, ymin=-1.0, ymax=1.0] {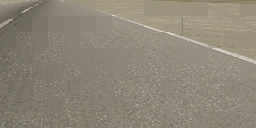};

\nextgroupplot [ylabel = {}, yticklabels={,,},xticklabels={,,}, enlargelimits = false, axis on top]
\addplot [point meta min=0, point meta max=50] graphics [xmin=-2.5, xmax=2.5, ymin=-1.0, ymax=1.0] {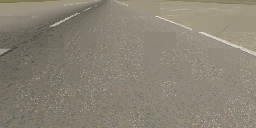};

\nextgroupplot [ylabel = {}, yticklabels={,,},xticklabels={,,}, enlargelimits = false, axis on top]
\addplot [point meta min=0, point meta max=50] graphics [xmin=-2.5, xmax=2.5, ymin=-1.0, ymax=1.0] {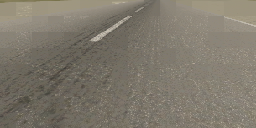};

\nextgroupplot [ylabel = {}, yticklabels={,,},xticklabels={,,}, enlargelimits = false, axis on top]
\addplot [point meta min=0, point meta max=50] graphics [xmin=-2.5, xmax=2.5, ymin=-1.0, ymax=1.0] {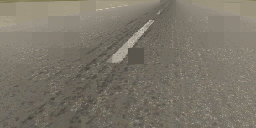};

\nextgroupplot [ylabel = {}, yticklabels={,,},xticklabels={,,}, enlargelimits = false, axis on top]
\addplot [point meta min=0, point meta max=50] graphics [xmin=-2.5, xmax=2.5, ymin=-1.0, ymax=1.0] {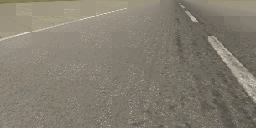};

\nextgroupplot [ylabel = {}, yticklabels={,,},xticklabels={,,}, enlargelimits = false, axis on top]
\addplot [point meta min=0, point meta max=50] graphics [xmin=-2.5, xmax=2.5, ymin=-1.0, ymax=1.0] {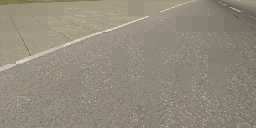};

\end{groupplot}
\end{tikzpicture}
    \caption{Images at different times during simulation of AST policy when using MCTS with two actions and $\delta_\text{max}=0.035$}
    \label{fig:Sequence}
\end{figure*}

\begin{table}
    \centering
    \caption{AST Results using DQN}
    \begin{tabular}{lccccc}  
    	\toprule
    	$N_\text{actions}$  & $\delta_\text{max}$ & Failure? & Steps & $\delta_\text{mean}$ & Log-likelihood \\
    	\midrule
    	2 & 0.035 & Yes & 28 & 0.035 & -197.23  \\
    	4 & 0.035 & Yes & 39 & 0.033 & -252.94  \\
    	6 & 0.035 & Yes & 27 & 0.0318 & -166.67 \\
    	\bottomrule
    \end{tabular}
    \label{tab:dqn}
\end{table}

Experimental results using DQN in AST instead of MCTS are shown in \Cref{tab:dqn}. DQN does not perform as well as MCTS for low numbers of actions but scales better to larger number of actions. Whereas MCTS uses only one initial state, DQN creates a policy that generalizes to any initial state, as shown in \Cref{fig:dqn}. DQN takes more samples to train the $Q$-network than to build a search tree, so MCTS required only 1--6 hours while DQN required 24--36 hours to run. However, both methods are model-free and return image disturbance sequences that lead the aircraft off the taxiway. When random samples are used instead of Marabou with $\delta_\text{max}=0.035$, the crosstrack position never exceeds $\SI{5}{\meter}$. This result demonstrates that using tools like Marabou are important for high dimensional problems.

\begin{figure}
    \centering
    \begin{tikzpicture}[]
\begin{groupplot}[height={3.5cm},width={\linewidth},group style={horizontal sep=0.0cm,vertical sep=1.0cm, group size=1 by 1}]

\nextgroupplot [ylabel={$d\ (\si{\meter})$},xlabel={Downtrack Position (m)}, enlargelimits = false, axis on top,xmin=-5.5,xmax=205.5,ymin=-10.7,ymax=13.5,legend pos={north west},
legend columns=2]]

\addplot graphics
[xmin=-5.5,xmax=205.5,ymin=-10.7,ymax=13.5]
{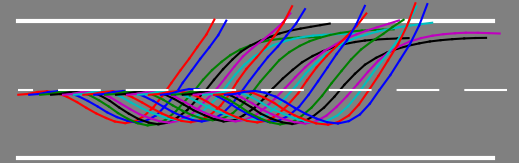};

\end{groupplot}

\end{tikzpicture}
    \caption{Simulations of AST policy from different initial downtrack positions}
    \label{fig:centerline}
\end{figure}

\Cref{fig:Sequence} shows four types of images at different times along the two action MCTS policy with $\delta_\text{max}=0.035$: the original taxiway image, downsampled image, downsampled image with perturbation added, and a reconstructed image of the taxiway image that would produce the perturbed image if downsampled. Reconstructing the taxiway image from the perturbed image is an under-defined problem, so this approach simply biases the brightness of pixels in the original taxiway image. The images reveal key insights into how the neural network can be tricked into failure. The reconstruction shows that AST darkens the taxiway edge lines until the thin boundary lines are difficult to discern, indicating that the boundary lines are important for making accurate predictions. 

Furthermore, times $t=14,15$ show that AST guides the aircraft across the centerline in the gap between centerline dashes. These images also have much greater $\epsilon$ values than previous times, which suggests that the neural network is more easily fooled when the aircraft travels between the centerline dashes. To investigate this hypothesis, the AST policy was simulated from different initial points along the taxiway. As shown in \Cref{fig:centerline}, trajectories that cross the centerline between dashes reach the edge of the taxiway while other trajectories remain on the taxiway.
\section{Conclusions}
Although neural networks perform well empirically, they can be susceptible to adversarial attacks. For safety critical image-based applications acting in the real world, verifying that failures will never occur is intractable or impossible. This work presented a method for validation of image-based neural network controllers that uses reinforcement learning to find the most likely failure modes. The analysis is tractable for high-dimensional inputs, and a taxiway navigation application demonstrated how the algorithm can be integrated with a black box simulator. The results showed that adaptive stress testing finds image disturbances that cause the neural network to guide the aircraft off the runway and revealed that the gaps between centerline dashes are more susceptible to adversarial perturbations.

Future work will study methods for making the neural network more robust to adversarial attacks and incorporate continuous actions. In addition, other image perturbations besides pixel disturbances could be considered, such as large skid marks or reflective puddles. Future work could also study how adding an additional camera to the left wing of the aircraft improves system robustness. Finally, a real aircraft on a taxiway could be used to validate the accuracy and applicability of the X-Plane 11 simulator.

\printbibliography
\end{document}